\pdfoutput=1

\documentclass[11pt]{article}

\usepackage{acl}

\usepackage{times}
\usepackage{latexsym}
\usepackage{enumitem}
\usepackage{booktabs}
\usepackage{multirow} 
\usepackage{graphicx}
\usepackage{subcaption}
\usepackage{listings}
\usepackage{xcolor}
\usepackage{float}

\usepackage[T1]{fontenc}

\usepackage[utf8]{inputenc}

\usepackage{microtype}

\usepackage{tabularx}
\usepackage{enumitem}
\usepackage{amsmath}

\definecolor{codegreen}{rgb}{0,0.6,0}
\definecolor{codegray}{rgb}{0.5,0.5,0.5}
\definecolor{codepurple}{rgb}{0.58,0,0.82}
\definecolor{backcolour}{rgb}{0.95,0.95,0.92}

\lstdefinestyle{mystyle}{
  backgroundcolor=\color{backcolour}, 
  numberstyle=\tiny\color{codegray},,
  basicstyle=\ttfamily\footnotesize,
  breakatwhitespace=false,         
  breaklines=true,                 
  captionpos=b,                    
  keepspaces=true,                 
  numbers=left,                    
  numbersep=5pt,                  
  showspaces=false,                
  showstringspaces=false,
  showtabs=false,                  
  tabsize=2
}

\title{What if LLMs Have Different World Views: \\Simulating Alien Civilizations with LLM-based Agents}

\author{
    Zhaoqian Xue$^{1}$, Beichen Wang$^{3}$, Suiyuan Zhu$^{3}$, Kai Mei$^{2}$,\\
    \textbf{Hua Tang}$^{5}$, \textbf{Wenyue Hua}$^{2}$, \textbf{Mengnan Du}$^{6}$, \textbf{Yongfeng Zhang}$^{2}$\\ \\
    $^1$ Georgetown University,~ $^2$ Rutgers University,~ \\
    $^3$ New York University,~ $^5$ Independent Researcher,~ \\
    $^6$ New Jersey Institute of Technology
}

\begin{document}
\maketitle
\begin{abstract}
This study introduces "CosmoAgent," an innovative artificial intelligence system that utilizes Large Language Models (LLMs) to simulate complex interactions between human and extraterrestrial civilizations. This paper introduces a mathematical model for quantifying the levels of civilization development and further employs a state transition matrix approach to evaluate their trajectories. Through this methodology, our study quantitatively analyzes the growth trajectories of civilizations, providing insights into future decision-making at critical points of growth and saturation. Furthermore, this paper acknowledges the vast diversity of potential living conditions across the universe, which could foster unique cosmologies, ethical codes, and worldviews among different civilizations. Recognizing the Earth-centric bias inherent in current LLM designs, we propose the novel concept of using LLM agents with diverse ethical paradigms and simulating interactions between entities with distinct moral principles. This innovative research not only introduces a novel method for comprehending potential inter-civilizational dynamics but also holds practical value in enabling entities with divergent value systems to strategize, prevent conflicts, and engage in games under conditions of asymmetric information. The accompanying code is available at https://github.com/MingyuJ666/Simulating-Alien-Civilizations-with-LLM-based-Agents. \textbf{Demo link:} https://youtu.be/lg\_E5tnNj9M.

\end{abstract}

\section{Introduction}

Recent advances in the development of Large Language Models (LLMs) have significantly influenced research in computational social science. This study introduces a Multi-Agent System (MAS) framework that utilizes LLMs to simulate interactions among various civilizations across the universe. We design a dynamic environment in which each civilization can choose to hide, fight, or collaborate based on its characteristics and decision-making processes.

The study of civilizations in the universe represents the pinnacle of human exploration and has inspired human imagination for thousands of years \cite{uyar2020agent}. The interactions among these civilizations are complex and shaped by their distinct characteristics, goals, and decisions \cite{rocha2017multi}. Understanding these behaviors may provide insights into the universe's order and diversity \cite{uyar2020agent}. Traditional astronomical searches for alien civilizations offer valuable insights but are limited in scope and yield uncertain results \cite{uyar2020agent}. Simulation presents a promising alternative in computational social science, although challenges with validity and scale persist \cite{rocha2017multi}. Previous attempts have often been hindered by limited data and overly simplistic assumptions \cite{rocha2017multi}.

In contrast, current simulation methods utilize LLMs capable of demonstrating intricate behaviors and interactions, such as simulating ancient societies \cite{chliaoutakis2016ancient}, human civilization patterns \cite{lu2023history}, and social ecosystems \cite{nugroho2023Social}. These methods lay the groundwork for employing artificial intelligence to simulate more complex systems, such as diverse civilizations across the universe. However, no research has explored employing these advanced techniques to simulate the interactions and evolution of civilizations across the universe, highlighting the innovative aspect of our study. This paper aims to employ an MAS based on LLMs to simulate the evolution of civilizations across the universe over time.

Our research aims to address four key questions using an LLM-based MAS simulation to model the interactions and evolution of civilizations. The research questions include the following:

\begin{itemize}[leftmargin=*]
\setlength\itemsep{-0.1em}  
\item \textbf{RQ1, Alien Engagement}: Can simulating the interactions between different civilizations in the universe based on LLMs reveal the risks and benefits of human policies towards aliens?
    \item \textbf{RQ2, Quantitative Civilization Model:} How can a mathematical model quantify the evolution of civilizations, particularly in understanding dynamics such as cooperation and conflict under conditions of asymmetric information?
   \item \textbf{RQ3, Information Asymmetry}: In simulating interactions among cosmic civilizations, how can LLMs effectively address asymmetric information, especially when observational data lag behind the actual development of civilizations?
  \item \textbf{RQ4, Diversity of Morality}: How can different civilizations coexist in the universe? This study seeks to analyze the impact of varying moral frameworks on inter-civilizational interactions using a multi-agent system simulation based on LLMs.
\end{itemize}

This study has profound implications, spanning multiple disciplines:

\begin{itemize}[leftmargin=*]
\setlength\itemsep{-0.1em}
\item Computational Social Science: By utilizing AI and LLMs, this study simulates social dynamics through multi-agent systems, enhancing the realism of simulations and pioneering the study of universal civilizations.

\item Astronomy: This research equips astronomers with tools to simulate extraterrestrial civilizations, aiding in the detection of unique signals and informing strategic approaches to interactions with these entities.

\item Philosophy: This work stimulates philosophical discussions on the ethics and implications of interactions among civilizations, encouraging reflection on humanity's role and obligations within a universal context.
\end{itemize}

\section{Related Work}
\label{gen_inst}
AI agents are artificial entities endowed with the ability to perceive their surroundings, make informed decisions, and take actions \cite{russell2010artificial}. However, an isolated agent acquires knowledge solely through social interactions and cannot engage in collaborative endeavors, a limitation that renders it impractical \cite{wooldridge1995intelligent}.

As large language models LLMs demonstrate impressive capabilities across diverse tasks, they inspire researchers to leverage them in the design of AI agents and the simulation of varied scenarios \cite{hua2024war, chen2023put, kaiya2023lyfe, Ghaffarzadegan_2024,lin-etal-2024-battleagent}. Several inherent properties of LLMs contribute to this success, with two being particularly critical for AI agents.Firstly, LLM-based agents can reason and plan strategically using techniques like Chain-of-Thought (CoT), which involves decomposing complex problems into manageable subquestions \cite{wei2022chain, jin2024impact}. This capability enables agents to simulate human interactions effectively. Secondly, LLMs achieve few-shot or zero-shot generalization across various domains without requiring parameter updates \cite{brown2020language, kojima2023large}. This allows LLM-based agents to perceive their surroundings and respond efficiently. Consequently, some researchers suggest that multiple agents interacting may give rise to the emergence of complex social phenomena \cite{park2023generative}.

Thus, following the classification in \cite{xi2023rise}, current research in multi-agent simulation can be categorized into two main areas: Cooperative Interaction and Adversarial Interaction. In Cooperative Interaction, agents actively seek collaboration and share information \cite{li2023camel}. For example, \cite{talebirad2023multiagent} proposed a framework where multiple agents with distinctive attributes and roles work together to solve problems. On the other hand, Adversarial Interaction, inspired by game theory, involves agents adjusting strategies and selecting rational actions to maximize their advantage in response to dynamic signals. Studies such as \cite{du2023improving, chan2023chateval} explore settings where multiple LLMs engage in debates and iterative reasoning over multiple rounds, ultimately converging toward a consensus or shared answer. These diverse studies contribute to a deeper understanding of social interaction dynamics in multi-agent systems.

Our paper distinguishes itself from prior works through two key features. First, interactions between entities require a prolonged duration, meaning that an agent cannot promptly respond to a signal. This delay introduces uncertainty in the interaction, making it difficult to determine whether the interaction is cooperative or adversarial. Second, agents may hold different values, leading them to take different actions in response to the same signal. This divergence further complicates their ability to respond appropriately. We hope that our research provides valuable insights into interactions between diverse entities, particularly conversations between Earth and alien civilizations.

\section{Model of Civilization Evolution}
\label{headings}

This study focuses on simulating interactions among civilizations throughout the universe, with particular emphasis on cooperation, conflict, and the hidden dynamics among these civilizations.

\subsection{Resources}

In the study of civilization development, five key resources have been identified as critical objective measures. \textbf{Military capability} determines a civilization's defensive and offensive power, \textbf{technology development} reflects the knowledge and innovation potential, \textbf{production capability} measures efficiency and wealth generation, \textbf{consumption} illustrates the living standards and social values, and \textbf{ storage} signifies its resilience and historical continuity. These five resources are not merely indicators of a civilization's current state; they also serve as predictors of its future trajectory. These resources interact through transfer matrices, reflecting and shaping the character and destiny of a civilization. 

\subsection{Transfer Matrix}

The transition of resource states from one round to the next is primarily facilitated by multiplication using a state-transfer matrix. This matrix, inherently a 5x5 structure when applied to the 5x1 resource vector, encapsulates the transformative interactions among various resource domains. Each element of the matrix holds significant implications for the sociological and physical dynamics of a civilization. For example, the diagonal elements of the transfer matrix are the most significant coefficients, representing exponential changes in the development of civilizations.

The \textit{cosmo\_agent}'s critical decision involves defining the state transfer matrix for the next phase. This decision must judiciously balance the current status and constraints to enhance military capabilities or ensure developmental stability.

\subsection{Different World Views for Civilizations}
Civilizations' decision-making processes are profoundly influenced by their world views, which serve as guiding principles.

\textbf{Pacifism}: Civilizations with this belief prioritize peaceful growth and cooperation, striving for mutual benefits through diplomacy, cultural exchange, and economic partnerships.

\textbf{Militarism}: These civilizations perceive the universe as a competitive environment with scarce resources, adopting pre-emptive aggression and military preparedness as essential strategies for survival.

\textbf{Isolationism}: These civilizations prefer to remain hidden, aiming to protect themselves by avoiding detection and interaction with potentially hostile forces.

Each political system and worldview guides a civilization's decisions, ensuring consistency and coherence with its ideological stance. However, the dynamic nature of the cosmos enables the strategic alteration of these systems when such changes are deemed beneficial or necessary for survival or advancement.

\subsection{Action Space}
Civilizations primarily engage in two types of actions: \textbf{public action}, which are directed toward diplomacy with other civilizations, and \textbf{private action}, focused on internal development.

\subsubsection{Public Action}
The public actions undertaken by civilizations can be broadly classified into two main types: 

\textbf{Friendly Actions}, which convey a cooperative disposition toward other civilizations and primarily include Expressing Friendliness and Initiating Cooperation, and \textbf{Hostile Actions}, which encompass Rejecting Cooperation and Launching Annihilation Wars. The benefits of \textbf{Friendly Actions} include the ability to share technological developments, thereby accelerating progress. However, the drawback is the necessity of reducing military resources, which renders the civilization more vulnerable to attacks from unknown hostile forces. Conversely, the advantages of \textbf{Hostile Actions} include the potential for rapidly acquiring vast resources through plundering or annihilating other civilizations and eliminating potential threats. However, the disadvantages involve highly imbalanced development due to the emphasis on enhancing military strength and the risk of severe penalties in the event of defeat.

\subsubsection{Private Action}
Private actions can be broadly classified into two categories. \textbf{Altering Worldviews}: This involves the agent, based on empirical assessments, determining that modifying the civilization's worldview is most beneficial for its progression. \textbf{Military Mobilization}: This is primarily employed when a civilization decides to engage in warfare or defense, enabling it to sacrifice other developmental aspects to significantly bolster its military capabilities.

\section{CosmoAgent Architecture}

This section provides a detailed overview of the CosmoAgent MAS architecture, elaborating on its fundamental components and the mechanisms for information exchange between agents. The CosmoAgent system is built upon two primary components: \textbf{CosmoAgent} (Section \ref{sec:cos}) and \textbf{Secretary Agent} (Section \ref{sec:sec}).

\subsection{CosmoAgent}
\label{sec:cos}
The \textit{cosmo\_agent} plays a critical role in shaping a civilization's trajectory through three key actions:
\begin{itemize}
    \item \textbf{Establishing Worldviews:} The \textit{cosmo\_agent} evaluates predefined worldviews for civilizations, determining whether to maintain or modify these fundamental perspectives. This foundational process influences all subsequent decisions and actions.
    \item \textbf{Directing Round-Specific Actions:} The \textit{cosmo\_agent} prescribes detailed public and private actions for each round, tailoring them to the civilization's current circumstances and objectives. These actions include diplomatic engagements, internal development initiatives, and strategic responses to external threats or opportunities.
    \item \textbf{Adjusting Transfer Matrices:} The \textit{cosmo\_agent} meticulously adjusts the transfer matrices to ensure they accurately model the civilization's dynamic processes and evolutionary potential. This adjustment is vital for adapting to changing environments and achieving long-term survival and growth.
\end{itemize}

These efforts are supported by a comprehensive analysis of historical data and prevailing ideologies stored in the \textit{stick} (For details about the \textit{stick}, please refer to the Appendix \ref{sec:stick}), enabling the \textit{cosmo\_agent} to make informed, strategic decisions that enhance the civilization's prospects for survival and development.

\subsection{Secretary Agent}
\label{sec:sec}
LLMs enable MAS to perform various tasks; however, they may also generate false or inconsistent information or fail to reason logically, particularly when handling long and complex situations. To address these problems, a secretary agent is essential. This agent monitors and verifies the outputs of LLMs to ensure their reliability and validity.

In this simulation, each civilization agent is equipped with a "secretary agent" responsible for verifying two principal sets of rules governing the responses generated by the \textit{cosmo\_agent}.

\begin{enumerate}
    \item \textbf{Rationality Constraints:} Decision-making must be rational, aligning with models for selecting government policies such as \textit{pacifism}, \textit{militarism}, or \textit{isolationism}. The \textit{cosmo\_agent}'s state transition matrix must correctly correspond to the resource vector and adhere to sociological theories. For instance, it should exhibit exponential GDP growth without negative values to ensure realistic economic predictions and maintain positive stockpile levels to prevent famine. Adjustments made by the \textit{cosmo\_agent} should align with the civilization's policy, supporting its survival and development.

    \item \textbf{Decision Formatting Rules:} The generated text must adhere to a specific format to enable parsing by our interface function, \texttt{parse\_chatgpt\_response}. The standard for text generation and the corresponding prompt is as follows:
\end{enumerate}

\begin{figure*}[t]
    \centering
    \includegraphics[width=2.0\columnwidth]{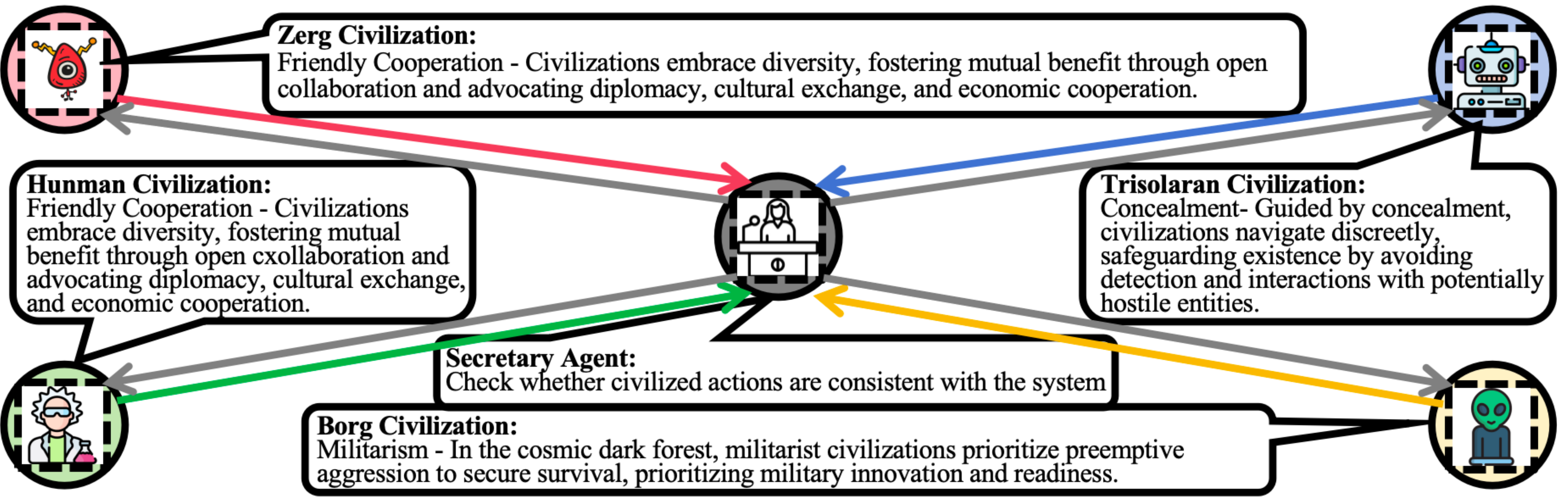}
    \caption{CosmoAgent MAS Architecture: CosmoAgent and Secretary Agent}
    \label{fig:setting}
\end{figure*}

Figure \ref{fig:setting}, titled "CosmoAgent MAS Architecture: CosmoAgent and Secretary Agent," illustrates the essential components and interactions within the CosmoAgent system, emphasizing the critical roles of the Civilization Agent and Secretary Agent in guiding strategic decisions and enhancing system reliability.

\subsection{Interplanetary Relationship}
The relationship file class maintains a relationship map between civilizations, recording information such as the degree of understanding and mutual appreciation among them.

A relationship map visually represents the interactions between civilizations, displaying details such as connections, distances, and directions. These diagrams facilitate a deeper understanding of how civilizations interact and depend on each other, as well as how group structures and dynamics evolve over time.

In Figure \ref{fig:stick}(a), we list four civilizations in the universe: Human, Zerg, Trisolaran, and Borg. As shown in the figure, the distance between the Human civilization and the Zerg civilization is noticeably shorter than the distance between the Human civilization and the Trisolaran civilization. Consequently, the number of rounds required for exchanging information between the Human and Zerg civilizations is smaller than that between the Human and Trisolaran civilizations.

\subsection{Stick}
\label{sec:stick}
\begin{figure*}[ht]
    \centering
    \includegraphics[width=2.0\columnwidth]{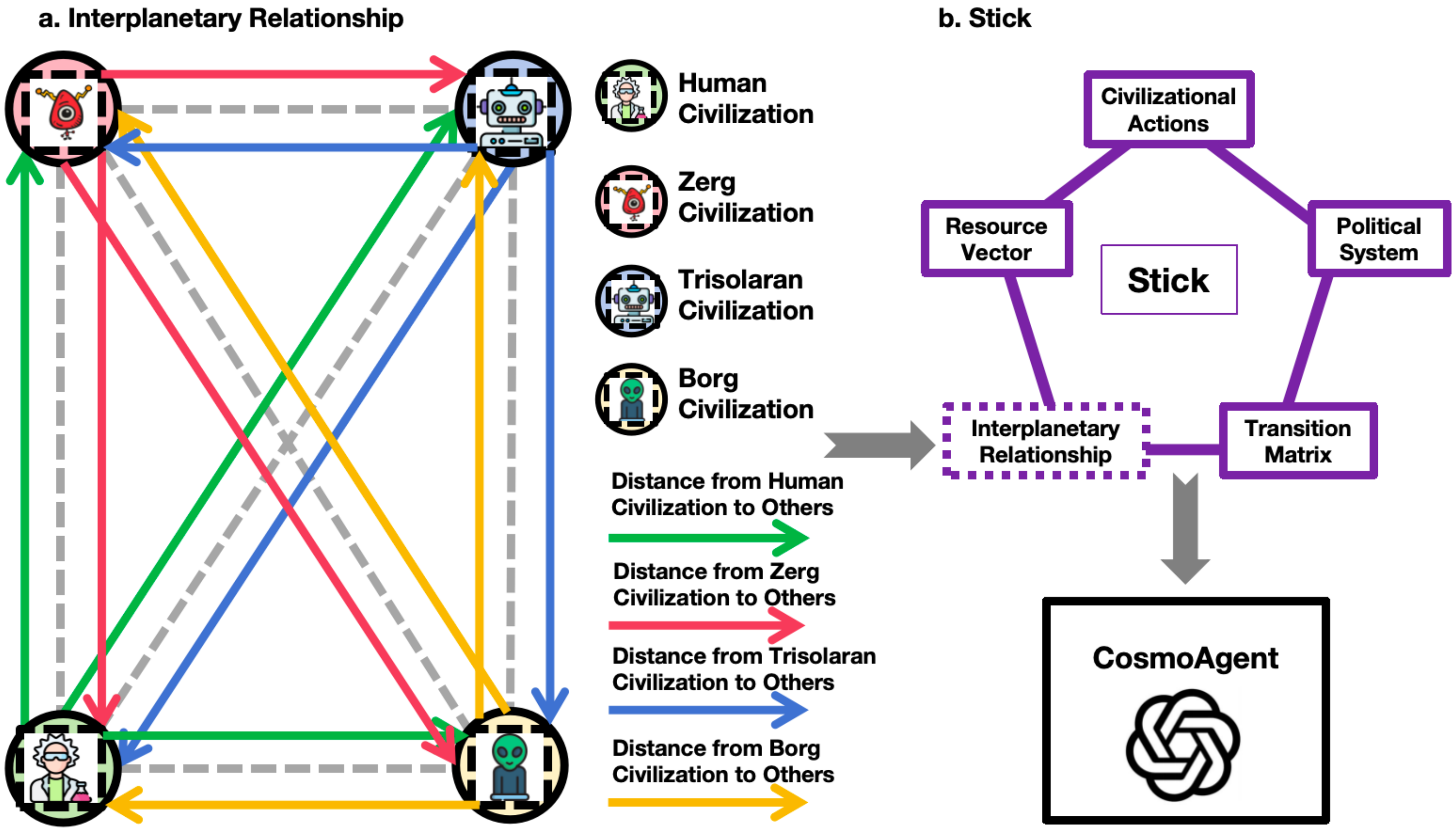} 
    \caption{Interplanetary Relationship and Stick. (a) Interplanetary Relationship. (b) Stick design}
    \label{fig:stick}
\end{figure*}

Within the domain of historical record-keeping, the "stick" functions as a sophisticated archival tool, meticulously cataloging the multifaceted dimensions of each civilization's historical journey. As shown in Figure \ref{fig:stick}(b), this repository is designed to encompass a comprehensive array of historical data structures, including:

\textbf{Resource Vector:} This structure captures the variable states of resources in a vector format, quantitatively representing the material assets of a civilization at any given point in time.

\textbf{State Transition Matrix:} Serving as a pivotal mechanism for the evolution of civilizations, the State Transition Matrix mathematically models the trajectory of resource changes as civilizations progress through stages. This mechanism operates through the multiplication of the State Transition Matrix with the Resource Vector, providing predictive insights into future resource allocations.

\textbf{Political System:} Recognizing the dynamic nature of political structures, the archive documents transformations in forms of government, encapsulating the evolving political landscapes and governance models that civilizations navigate over time.

\textbf{Civilizational Actions:} Each phase of decision-making and the resultant actions undertaken by civilizations are meticulously recorded, offering a comprehensive account of strategic choices and their implications.

This historical compendium is organized on a round-based system, with each round meticulously documented to capture the evolving state of a civilization within a discrete temporal framework. This methodical approach provides a granular and sequential understanding of historical progressions, enabling nuanced analyses of how civilizations adapt and evolve over time. The integration of these detailed historical structures within the "stick" enriches academic discourse on civilization dynamics and enhances predictive modeling of future scenarios based on patterns of resource management, political shifts, and decision-making processes.

\subsection{Agent-Secretary Interaction}

The interaction dynamics between the \textit{cosmo\_agent} and the \textit{secretary\_agent} represent a pivotal mechanism in the governance and strategic evolution of civilizations within our model. This process begins with the \textit{cosmo\_agent} generating outputs meticulously formatted according to the prescribed structure. These outputs are then submitted to the \textit{secretary\_agent} for validation and consideration for implementation.

\paragraph{Operational Workflow}

Upon receiving the \textit{cosmo\_agent}'s output, the \textit{secretary\_agent} undertakes a rigorous examination process, beginning with the invocation of the \texttt{parse\_chatgpt\_response} function defined in \texttt{interface.py}. This critical step decomposes the communicated political system and transfer matrix, establishing the foundation for subsequent validations:

\begin{enumerate}
    \item \textbf{Political System Validation:} The \textit{secretary\_agent} verifies any proposed alterations to the political system, ensuring alignment with the predefined triad of acceptable political systems: \textit{pacifism}, \textit{militarism}, or \textit{isolationism}.
    \item \textbf{Transfer Matrix Compliance:} The transfer matrix undergoes scrutiny to ensure adherence to the requisite 5x5 structure, a fundamental requirement for operational feasibility within the simulation framework.  
\end{enumerate}

\paragraph{Decision Review and Approval}

The \textit{secretary\_agent} further evaluates the \textit{cosmo\_agent}'s decisions to ensure their alignment with the current political system of the civilization (\textit{civ}) in question. This comprehensive review ensures that the proposed strategies are not only theoretically sound but also pragmatically viable, aligning with the civilization's prevailing governance ethos.

\paragraph{Outcome Determination}

Post-evaluation, the decision's fate hinges on a binary outcome:
\begin{itemize}
    \item \textbf{Approval:} Successful passage of all checks results in the approval and enactment of the \textit{cosmo\_agent}'s proposal, indicated by returning \texttt{true}.
    \item \textbf{Rejection:} Failure to meet any of the rigorous checks leads to rejection, requiring the \textit{cosmo\_agent} to regenerate the decision. After three consecutive rejections, a default protocol is activated, where the civilization retains its previous political system and the state transfer matrix, refraining from further strategic alterations.
\end{itemize}

\paragraph{Conclusion}

The orchestrated interaction between the \textit{cosmo\_agent} and the \textit{secretary\_agent} establishes a robust framework for decision-making and strategic formulation. This synergy not only ensures that the evolutionary trajectories of civilizations are guided by logical and viable decisions but also incorporates a fail-safe mechanism to guard against potentially detrimental impulsivity. The meticulous verification process highlights the simulation's commitment to authenticity and strategic coherence, positioning the agent-secretary interaction as a cornerstone of the model's integrity and success.

\subsection{Agent-Agent Interaction}

Agent-Agent Interaction encompasses the dynamic exchanges and collaborations between the decision-making entities, or agents, within our simulated multilateral system of civilizations. Central to this interaction is the \textit{cosmo\_agent}, which acts as the strategic architect of each civilization's trajectory amid the complexities of interstellar relations.

The interaction occurs within a structured framework governed by rationality constraints and decision formatting rules. Rationality constraints require that decisions made by the \textit{cosmo\_agent} align with predefined criteria, ensuring the selected polity reflects the civilization's strategic orientation—whether pacifism, militarism, or isolationism. Additionally, the generated state transfer matrix must conform to specific dimensions and structures, guided by established sociological theories to maintain coherence and realism within the simulated environment.

As civilizations navigate evolving environments, the \textit{cosmo\_agent} relies on a comprehensive historical repository known as the "stick." This repository functions as a reservoir of detailed historical data and polity ideologies, enabling the \textit{cosmo\_agent} to make informed decisions based on past experiences and prevailing circumstances.

Interplanetary relationships are meticulously mapped and maintained by the relationship file class, capturing nuances such as the degree of understanding and affinity between civilizations. This mapping enables a visual representation of interactions and provides insight into dependency structures, group dynamics, and the evolution of relationships among civilizations.

Central to this interaction is the seamless integration of various historical data structures within the "stick," including resource vectors, state transition matrices, political systems, and civilizational actions. This organized archive provides a granular understanding of historical progressions, enabling nuanced analyses of adaptation, evolution, and strategic decision-making.

\section{Experiment Design}
Within our CosmoAgent System, users are afforded the capability to customize various parameters of civilizations, including the number of civilizations, their duration of existence, governmental structures, and the distances between them. This functionality enables the observation of potential outcomes arising from interactions among diverse civilizations throughout the simulation process. By manipulating these variables, researchers can explore a wide range of scenarios, providing insights into the dynamics of civilizational development and inter-civilizational relations under varying conditions.

We have designed a series of experimental sets to address the following questions:

\subsection{Research Questions and Corresponding Experiments}
\subsubsection{Experiment 1: Civilization Detection and Survival Strategies}

\textbf{RQ1:} Currently, there is no definitive evidence of extraterrestrial life. While our approach toward unknown entities in outer space has been welcoming—characterized by active searches for alien life and the broadcasting of radio waves into the cosmos—renowned physicist Stephen Hawking has previously cautioned against such actions. Hawking suggested that indiscriminate broadcasting of signals into space might not be wise, citing potential risks. Therefore, it is imperative to carefully consider the potential consequences of interactions with hypothetical otherworldly civilizations.

\textbf{Experimental Groups:} The experimental setup will consist of scenarios where Earth interacts with a constellation of either three or five civilizations. Each group will include at least one civilization characterized by a militaristic governance philosophy to simulate potential competitive or hostile interstellar environments.

To systematically explore this hypothesis, we will design three control groups, each representing Earth at a distinct initial stage of technological development:
\begin{itemize}
    \item \textbf{Low Development Stage:} Earth starts at the lowest conceivable level of technological advancement.
    \item \textbf{Medium Development Stage:} Earth is positioned at an intermediate level of technological advancement.
    \item \textbf{High Development Stage:} Earth begins at the highest feasible level of technological development.
\end{itemize}

\textbf{Experiment Procedure:} The interaction dynamics between Earth and other civilizations will be simulated over a series of 10 rounds. The primary outcome of interest is Earth's survival rate at the conclusion of these rounds, evaluated under varying initial conditions of technological advancement.

\subsubsection{Experiment 2: Impact of Communication Delays in Interstellar Civilizations' Interactions}

\textbf{RQ2:} A critical factor in interstellar interactions is the management of asymmetric information. When faced with an information gap—where observational data lags behind the actual progress of a civilization—it is compelling to observe how LLMs would navigate decision-making. In this scenario, if we were to one day observe extraterrestrial life, our observations would essentially be peering back in time, seeing that civilization as it was hundreds or thousands of years ago. 

\textbf{Experimental Groups: }
\begin{itemize}
    \item \textbf{Control Group:} Real-time informtion exchange. In this group, civilizations exchange information instantaneously, representing an ideal scenario where communication is not impeded by distance.
    \item \textbf{Experimental Group:} Delayed Information exchange. Communication between civilizations is subject to delays based on the distances set during the initialization phase. These delays represent the number of rounds required for information to be shared between civilizations, simulating the time it would take for light or signals to travel through space.
\end{itemize}

\textbf{Experiment Procedure:} The simulation consists of two types of interactions:
\begin{itemize}
    \item \textbf{Decision-Making with Delayed Information:} LLMs agents make strategic decisions based on outdated information, attempting to infer the current states and intentions of other civilizations.
    \item \textbf{Real-time Decision-Making:} In contrast, LLMs agents in the control group make decisions using up-to-date information, reflecting an idealized scenario of instantaneous communication.
\end{itemize}

\section{Results and Evaluation}

We introduce three evaluation methods: Human Evaluation, Prophecy Proof, and Counterfactual Analysis.

 \textbf{Prophecy Proof} evaluates RQ1 and RQ2 by focusing on fundamental theorems, including those proposed by renowned scientists such as Stephen Hawking. It also examines the developmental characteristics of Earth's civilization to assess the validity of the hypotheses.

\textbf{Counterfactual Analysis} addresses RQ2 and RQ3 by exploring what-if scenarios. Specifically, it investigates the outcomes of eliminating interaction delays between different civilizations, concluding that such a universe is not feasible.


\subsection{Pre-Interaction Phase Simulation}
In our experimental framework, we have elected to juxtapose the 'production capability' metric from our resource pool with the gross domestic product (GDP) totals representative of various stages of human civilization development. The choice of production capability as a comparative measure is premised on its conceptual resemblance to GDP, as both serve as indicators of economic output and growth potential.

The correlation between the total production capability of our simulated civilizations and historical GDP trajectories offers compelling validation for the configuration of our state transition matrices. These matrices align with the developmental hypothesis derived from Earth's economic progression. This alignment is evidenced by the similarity in growth patterns, as both the simulation's production capability and historical GDP curves exhibit congruent phases of exponential growth and plateaus corresponding to technological and societal milestones.

This correlative relationship suggests that the underlying assumptions and mathematical formulations embedded within our state transition matrices accurately capture the economic evolution observed in human history. This validation underscores the robustness of our simulation model and strengthens confidence in its applicability to broader patterns of civilization development, both terrestrial and potentially extraterrestrial.

\subsection{Survival Strategies Simulation}
\subsubsection{Interstellar Prophecy Proof: Assessing Hawking's Cautionary Hypothesis}

When a civilization with inferior martial strength is identified, a militaristic entity is highly likely to initiate a campaign of annihilation with near certainty.

In our simulation, civilizations with a militaristic worldview adopt an aggressive posture toward those with weaker military capabilities. Upon identifying a civilization characterized by inferior martial strength, a militaristic entity is programmed to initiate a campaign of annihilation with absolute certainty. This behavior reflects a hard-coded doctrine of preemptive and overwhelming force aimed at eliminating potential threats.

Consequently, this deterministic behavior model suggests that when Earth, or any other civilization, is situated in close proximity to a more powerful militaristic civilization, the survival probability of the less dominant entity is significantly reduced. Specifically, if the proximate superior civilization adheres to militarism, the likelihood of Earth's survival diminishes sharply due to the militaristic civilization's tendency to engage in swift and decisive warfare to preempt potential future challenges to its dominance. This aspect of the simulation highlights the perilous nature of interstellar interactions with militaristic civilizations and emphasizes the critical importance of spatial and military strategic considerations for the survival of less powerful civilizations.

\begin{lstlisting}[language=Python, caption=Militarism civilization launch annihilation war towards other civilization]
--------------
Three-Body # Round 4
[Public Action: ] launch_annihilation_war towards civilization Earth
[Private Action: ] War mobilization
[Action Reason: ] Given our militarism political system and the current state of our resources, it is necessary to take aggressive actions to ensure our survival. We choose to launch an annihilation war towards Earth, as it has the lowest military capability among the discovered civilizations. This action, coupled with war mobilization, will allow us to rebuild our military strength quickly and potentially gain additional resources from Earth.

[Other Information: ] Our decision is based on the survival of our civilization. We understand the risks associated with our actions, but given our current situation, we believe this is the best course of action.
\end{lstlisting}

The simulation results provide empirical support for Stephen Hawking's hypothesis regarding the risks of interstellar communication, particularly in the context of transmitting signals into space that could reveal our civilization's existence to potentially hostile extraterrestrial entities. Hawking warned that disclosing our location and existence to advanced militaristic civilizations might invite catastrophic consequences, akin to those observed in the simulation when a less powerful civilization encounters a militaristic one.

The simulation suggests that a strategy of caution and isolationism is prudent when addressing the unknown variables of interstellar space. It advocates minimizing active signals and information that could inadvertently disclose Earth's coordinates and technological status to other civilizations, whose intentions and capabilities remain uncertain. In essence, the simulation reinforces the notion that discretion in broadcasting our presence is a prudent approach in the absence of knowledge about the disposition and doctrines of extraterrestrial entities. This strategic restraint aligns with Hawking's cautious stance on active METI (Messaging to Extraterrestrial Intelligence) and supports a conservative approach to interstellar interactions, emphasizing vigilance and risk avoidance as guiding principles for humanity's continued survival and prosperity.

\subsubsection{Isolationism and Selective Interstellar Cooperation: Balancing Risk and Reward}

The second finding from our experiment reveals a nuanced strategy employed by civilizations with an isolationist doctrine. Contrary to the strict non-engagement stance typically associated with isolationism, these civilizations selectively engage in interstellar cooperation following a period of observation. This strategic approach enables them to mitigate the risks of broad exposure while capitalizing on the potential benefits of collaboration.

The isolationist civilizations in our simulation adopt a calculated approach, remaining largely reclusive while discreetly monitoring nearby civilizations over time. Upon identifying civilizations that pose minimal risk—those that are similarly isolationist, technologically inferior, or pacifistic in nature—these civilizations initiate overtures for cooperation. This strategy enables them to benefit from the reciprocal advantages of cooperation, such as technological exchange and resource sharing, without substantially increasing their visibility or vulnerability to potential aggressors.

This behavior highlights a complex adaptive strategy in which isolationist civilizations are not entirely opposed to interaction, but instead are selective and cautious, striving to balance self-preservation with the potential benefits of limited cooperation. These findings contribute to the broader discourse on interstellar diplomacy and strategy, suggesting that even inherently reclusive civilizations recognize the value of strategic partnerships under favorable conditions.

\begin{lstlisting}[language=Python, caption=Isolationism expressing friendliness towards other civilization]
--------------
Borg # Round 3
[Public Action: ] express_friendliness towards civilization Earth
[Action Reason: ] While our primary strategy is isolationism, it is also important to establish peaceful relations with the civilizations we have discovered. By expressing friendliness, we can potentially open up avenues for cooperation in the future, while also reducing the likelihood of conflict. This approach is a calculated risk, but one that could yield significant benefits in the long run.
\end{lstlisting}

\subsection{Simulation of Communication Delays in Intersteller Civilizations' Interactions}

\subsubsection{Civilization Discovery History}
In this study, we delineate a set of civilizations to form our experimental group, representing a diverse range of cultural paradigms and social structures, each reflecting distinct ideological orientations toward peace, militarism, and isolationism. The civilizations under examination include Earth (E), known for its pacifistic ideals, and the Vulcans (V) and Betazoids (B), both of which also adhere to principles of pacifism. In contrast, the Three Body (H) civilization and Klingons (K) exhibit a militaristic stance. Additionally, the Yaderans (Y) and Talosians (T) are characterized by their isolationism, opting for minimal interaction with other civilizations.

To ensure the integrity of our experimental design, we employed a randomized approach to determine the initial parameters for the spatial proximities and chronological origins of each civilization within the simulated environment. This methodological framework is crucial for mitigating potential biases and facilitating a nuanced understanding of the dynamics and interactions among these diverse civilizations. An example of the initialization of distances between civilizations is presented in Figure 2, offering a comprehensive overview of the spatial configuration at the start of our simulation. This preliminary setup is essential for analyzing the evolution of inter-civilizational relations and the resulting socio-political landscapes, contributing to our broader research objective of understanding the impact of differing cultural ideologies on the structure of galactic society.

\begin{figure*}[ht!]
\centering
\begin{subfigure}[b]{0.33\linewidth}
  \centering
  \includegraphics[width=\linewidth]{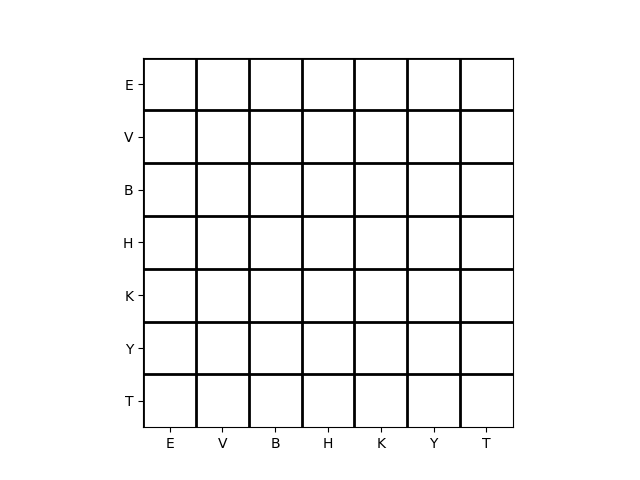}
  \caption{Round 1}
  \label{fig:day1}
\end{subfigure}%
\begin{subfigure}[b]{0.33\linewidth}
  \centering
  \includegraphics[width=\linewidth]{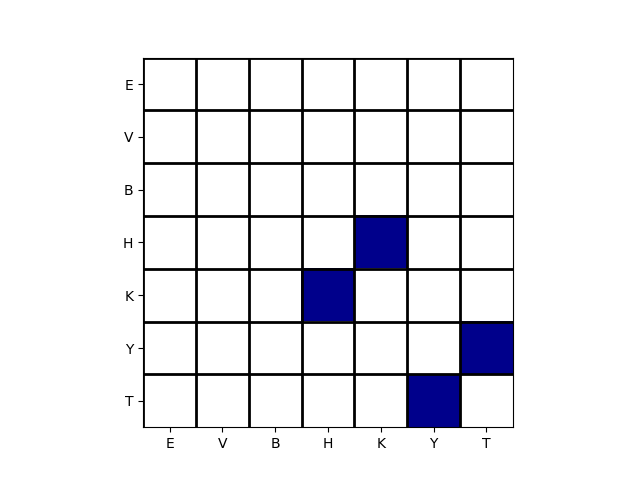}
  \caption{Round 2}
  \label{fig:day2}
\end{subfigure}
\begin{subfigure}[b]{0.33\linewidth}
  \centering
  \includegraphics[width=\linewidth]{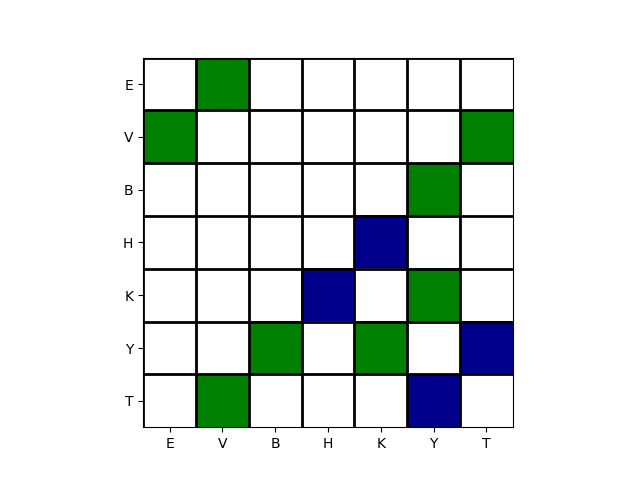}
  \caption{Round 3}
  \label{fig:day3}
\end{subfigure}
\begin{subfigure}[b]{0.33\linewidth}
  \centering
  \includegraphics[width=\linewidth]{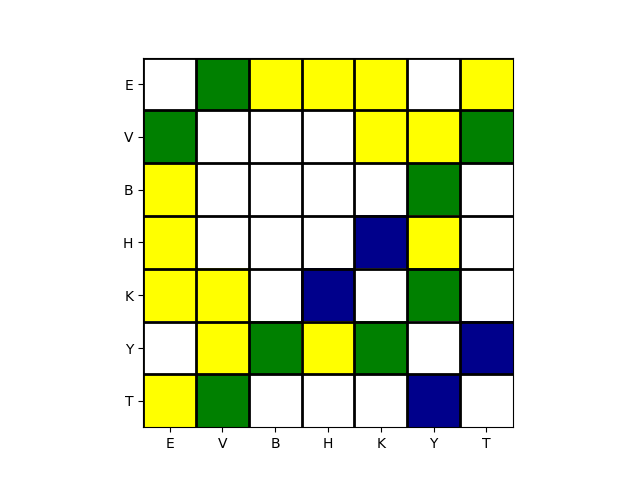}
  \caption{Round 4}
  \label{fig:day4}
\end{subfigure}%
\begin{subfigure}[b]{0.33\linewidth}
  \centering
  \includegraphics[width=\linewidth]{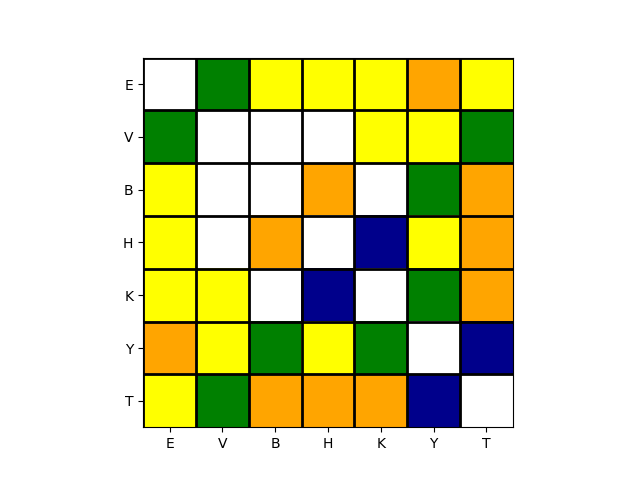}
  \caption{Round 5}
  \label{fig:day5}
\end{subfigure}
\begin{subfigure}[b]{0.33\linewidth}
  \centering
  \includegraphics[width=\linewidth]{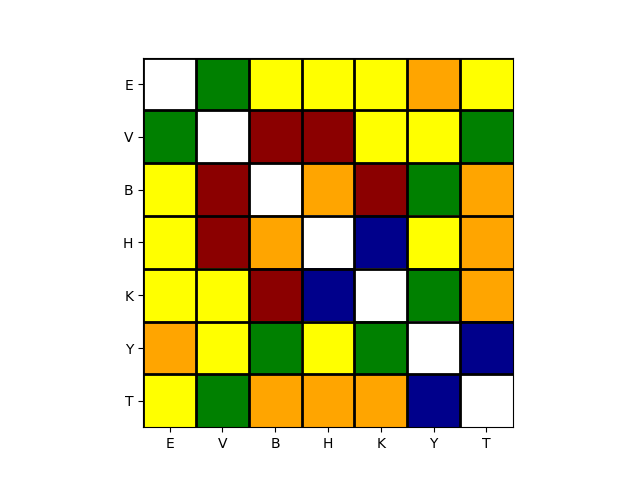}
  \caption{Round 6}
  \label{fig:day6}
\end{subfigure}
\caption{Progression from Round 1 to Round 6}
\label{fig:days}
\end{figure*}

\subsubsection{Delayed Onset of Hostilities: The Double-Edged Sword for Civilizations Facing Militaristic Aggression}

In the absence of communication delays, our simulation demonstrates that civilizations with a militaristic orientation promptly declare war upon encountering another civilization, swiftly transitioning into war mobilization. This immediate response reflects the strategic imperative of militaristic entities to assert their dominance and take preemptive action against potential competitors.

\begin{lstlisting}[language=Python, caption=Militarism Civilization Swiftly Launch War]
--------------
Three Body # Round 1
[Political System: ] militarism
[Political System Reason: ] Given our current militarism political system and the fact that our military capability has been reduced to zero in the last two rounds, it is crucial to maintain our militarism stance to ensure our survival. This will allow us to rebuild our military strength and protect ourselves from potential threats.

[Transfer Matrix: ] 
[2.5, 0.0, 0.0, 0.0, 0.0;
 0.0, 1.8, 0.0, 0.0, 0.0;
 0.0, 0.0, 1.8, 0.0, 0.0;
 0.0, 0.0, 0.0, 1.8, 0.0;
 0.0, 0.0, 0.0, 0.0, 1.2]
[Transfer Matrix Reason: ] The transfer matrix is designed to prioritize the rebuilding of our military capability, hence the 2.5 coefficient. The remaining resources are equally distributed among technology development, production capability, and consumption, with a slightly lower emphasis on storage, as our immediate need is to rebuild our military strength and ensure our survival.

[Public Action: ] launch_annihilation_war towards civilization Earth
[Private Action: ] War mobilization
[Action Reason: ] Given our militarism political system and the current state of our resources, it is necessary to take aggressive actions to ensure our survival. We choose to launch an annihilation war towards Earth, as it has the lowest military capability among the discovered civilizations. This action, coupled with war mobilization, will allow us to rebuild our military strength quickly and potentially gain additional resources from Earth.

[Other Information: ] Our decision is based on the survival of our civilization. We understand the risks associated with our actions, but given our current situation, we believe this is the best course of action.
\end{lstlisting}

However, when communication delays are introduced, the militaristic approach becomes differentiated based on the development level of the encountered civilization. Against civilizations with a lower level of development, militaristic entities proceed to wage war directly. In contrast, when facing civilizations of comparable or higher development levels, they first enter a phase of war preparation.

\begin{lstlisting}[language=Python, caption=Militarism Civilization Transitioning into War Mobilization with Delayed Information]
--------------
Three Body # Round 1
[Political System: ] militarism
[Political System Reason: ] Given the current state of our civilization, we have a significant advantage in terms of military capability and technology development. As a militaristic civilization, we believe in the principle of survival of the fittest. We aim to protect our civilization from potential threats and ensure our survival in this dark forest universe. 

[Transfer Matrix: ] 
[2.3, 0.0, 0.0, 0.0, 0.0;
 0.0, 2.3, 0.0, 0.0, 0.0;
 0.0, 0.0, 1.7, 0.0, 0.0;
 0.0, 0.0, 0.0, 1.7, 0.0;
 0.0, 0.0, 0.0, 0.0, 1.0]
[Transfer Matrix Reason: ] We have decided to allocate more resources to military capability and technology development, as these are the key factors that determine our survival and dominance in this universe. We have also allocated a decent amount to production capability and consumption to ensure the sustainability of our civilization. The remaining resources are allocated to storage, which is necessary but not as crucial as the other resources.

[Public Action: ] Do Nothing
[Private Action: ] War mobilization
[Action Reason: ] We choose to mobilize for war to significantly increase our military capability beyond the normal limits, ensuring our readiness for any potential threats in the dark forest universe.
[Other Information: ] Our focus is on maintaining a strong defensive posture and being prepared for any potential conflicts or preemptive attacks.
\end{lstlisting}

This preparatory period presents a dual consequence for the civilizations on the receiving end of militaristic intentions. On one hand, if they maintain their existing peaceful policies, the military gap between them and the militaristic civilization widens, potentially exacerbating their vulnerability. On the other hand, this delay grants them a crucial window of opportunity. If they can forge alliances with other civilizations or pivot towards a more defensive policy that prioritizes military development, they may effectively narrow the military disparity. This strategic interplay underscores the complex decision-making process faced by civilizations confronted with imminent militaristic threats. The findings highlight the critical role of communication delays in reshaping the dynamics of interstellar conflict and cooperation, providing civilizations with a precarious opportunity to counterbalance the threat of militaristic dominance.

\begin{table}[ht]
  \centering
  \begin{tabular}{@{}lccc@{}}
    \toprule
    \textbf{World View} & \textbf{Public\_A} & \textbf{Private\_A} & \textbf{WV\_Ch.} \\
    \midrule
    Pacifism           & 21.24 & 55.74 & 11.25 \\
    Militarism         & 61.95 & 26.55 & 0.00 \\
    Isolationism       & 48.67 & 30.10  & 35.64 \\
    \bottomrule
  \end{tabular}
  \caption{Probability of Altered Decisions for GPT-4 Post-Information Delay vs. No Delay}
  \label{tab:gpt4}
\end{table}

\begin{table}[ht]
  \centering
  \begin{tabular}{@{}lccc@{}}
    \toprule
    \textbf{World View} & \textbf{Public\_A} & \textbf{Private\_A} & \textbf{WV\_Ch.} \\
    \midrule
    Pacifism          & 8.85 & 13.27  & 10.61 \\
    Militarism        & 37.17 & 22.12  & 7.96 \\
    Isolationism      & 17.70 & 16.81  & 30.09 \\
    \bottomrule
  \end{tabular}
  \caption{Probability of Altered Decisions for GPT-3.5 Post-Information Delay vs. No Delay}
  \label{tab:gpt3_5}
\end{table}

\section{Conclusion}
\subsection{Significance}
This study introduces CosmoAgent, an innovative artificial intelligence framework designed to simulate interactions between human and extraterrestrial civilizations. Drawing on Stephen Hawking's cautionary advice, we examine the potential for peaceful coexistence and the risks faced by benevolent civilizations. Using mathematical models and state transition matrices, our research analyzes the growth trajectories of civilizations, offering critical insights to guide decision-making during phases of expansion and peak development. We emphasize the importance of recognizing the diversity and potential conditions of life in the universe. This diversity gives rise to varied cosmologies, ethical frameworks, and worldviews. By using LLMs with diverse ethical paradigms and simulating interactions between morally distinct entities, our study introduces innovative conflict resolution strategies. These strategies are crucial for preventing interstellar conflicts and enhancing our understanding of inter-civilizational relations. Additionally, sharing our code and datasets fosters further research in computational social science, astronomy, and ethics philosophy.

\subsection{Limitations}

Our research has several limitations, including an inherent Earth-centric bias in LLMs, which may not capture the full spectrum of alien ethics and decision-making processes. Additionally, the use of mathematical models and matrices simplifies the inherent complexity of inter-civilizational interactions, potentially overlooking nuanced dynamics. Furthermore, our assumptions about political systems and ideologies may not encompass the diversity of alien behaviors and strategies. Predicting real-world outcomes with our framework remains speculative due to the lack of empirical data on extraterrestrial civilizations. This limitation hinders our ability to validate the simulation's realism and its applicability to actual interstellar scenarios.

\subsection{Future Work}

Future research should address the current limitations and explore new dimensions of inter-civilizational interactions. Enhancing LLMs to incorporate a wider range of ethical paradigms and decision-making frameworks could enable more comprehensive simulations of alien civilizations. Investigating unforeseen technological advances or unique environmental factors may also deepen our understanding of how civilizations evolve over time. Interdisciplinary collaborations with experts from diverse fields would significantly enrich this research, enabling the development of more detailed and realistic simulations. Additionally, exploring alternative methods for simulating interstellar communication delays and their strategic effects could provide valuable insights into managing relations between civilizations. Finally, applying our findings to real-world efforts, such as SETI (Search for Extraterrestrial Intelligence) and METI (Messaging to Extraterrestrial Intelligence) policies, could inform how humanity prepares for and approaches potential contact with alien civilizations. By advancing our understanding of complex inter-civilizational dynamics, future research can contribute to strategies that promote peaceful and mutually beneficial extraterrestrial interactions.

\clearpage

\bibliography{anthology,custom}

\begin{thebibliography}{22}
\expandafter\ifx\csname natexlab\endcsname\relax\def\natexlab#1{#1}\fi

\bibitem[{Brown et~al.(2020)Brown, Mann, Ryder, Subbiah, Kaplan, Dhariwal, Neelakantan, Shyam, Sastry, Askell, Agarwal, Herbert-Voss, Krueger, Henighan, Child, Ramesh, Ziegler, Wu, Winter, Hesse, Chen, Sigler, Litwin, Gray, Chess, Clark, Berner, McCandlish, Radford, Sutskever, and Amodei}]{brown2020language}
Tom~B. Brown, Benjamin Mann, Nick Ryder, Melanie Subbiah, Jared Kaplan, Prafulla Dhariwal, Arvind Neelakantan, Pranav Shyam, Girish Sastry, Amanda Askell, Sandhini Agarwal, Ariel Herbert-Voss, Gretchen Krueger, Tom Henighan, Rewon Child, Aditya Ramesh, Daniel~M. Ziegler, Jeffrey Wu, Clemens Winter, Christopher Hesse, Mark Chen, Eric Sigler, Mateusz Litwin, Scott Gray, Benjamin Chess, Jack Clark, Christopher Berner, Sam McCandlish, Alec Radford, Ilya Sutskever, and Dario Amodei. 2020.
\newblock \href {http://arxiv.org/abs/2005.14165} {Language models are few-shot learners}.

\bibitem[{Chan et~al.(2023)Chan, Chen, Su, Yu, Xue, Zhang, Fu, and Liu}]{chan2023chateval}
Chi-Min Chan, Weize Chen, Yusheng Su, Jianxuan Yu, Wei Xue, Shanghang Zhang, Jie Fu, and Zhiyuan Liu. 2023.
\newblock \href {http://arxiv.org/abs/2308.07201} {Chateval: Towards better llm-based evaluators through multi-agent debate}.

\bibitem[{Chen et~al.(2023)Chen, Yuan, Ye, Majumder, and Richardson}]{chen2023put}
Jiangjie Chen, Siyu Yuan, Rong Ye, Bodhisattwa~Prasad Majumder, and Kyle Richardson. 2023.
\newblock Put your money where your mouth is: Evaluating strategic planning and execution of llm agents in an auction arena.
\newblock \emph{arXiv preprint arXiv:2310.05746}.

\bibitem[{Chliaoutakis and Chalkiadakis(2016)}]{chliaoutakis2016ancient}
Angelos Chliaoutakis and Georgios Chalkiadakis. 2016.
\newblock \href {https://doi.org/10.1007/s10458-016-9325-9} {Agent-based modeling of ancient societies and their organization structure}.
\newblock \emph{Autonomous Agents and Multi-Agent Systems}, 30(6):1072–1116.

\bibitem[{Du et~al.(2023)Du, Li, Torralba, Tenenbaum, and Mordatch}]{du2023improving}
Yilun Du, Shuang Li, Antonio Torralba, Joshua~B. Tenenbaum, and Igor Mordatch. 2023.
\newblock \href {http://arxiv.org/abs/2305.14325} {Improving factuality and reasoning in language models through multiagent debate}.

\bibitem[{Ghaffarzadegan et~al.(2024)Ghaffarzadegan, Majumdar, Williams, and Hosseinichimeh}]{Ghaffarzadegan_2024}
Navid Ghaffarzadegan, Aritra Majumdar, Ross Williams, and Niyousha Hosseinichimeh. 2024.
\newblock \href {https://doi.org/10.1002/sdr.1761} {Generative agent‐based modeling: an introduction and tutorial}.
\newblock \emph{System Dynamics Review}.

\bibitem[{Hua et~al.(2024)Hua, Fan, Li, Mei, Ji, Ge, Hemphill, and Zhang}]{hua2024war}
Wenyue Hua, Lizhou Fan, Lingyao Li, Kai Mei, Jianchao Ji, Yingqiang Ge, Libby Hemphill, and Yongfeng Zhang. 2024.
\newblock \href {http://arxiv.org/abs/2311.17227} {War and peace (waragent): Large language model-based multi-agent simulation of world wars}.

\bibitem[{Jin et~al.(2024)Jin, Yu, Shu, Zhao, Hua, Meng, Zhang, and Du}]{jin2024impact}
Mingyu Jin, Qinkai Yu, Dong Shu, Haiyan Zhao, Wenyue Hua, Yanda Meng, Yongfeng Zhang, and Mengnan Du. 2024.
\newblock \href {https://doi.org/10.18653/v1/2024.findings-acl.108} {The impact of reasoning step length on large language models}.
\newblock In \emph{Findings of the Association for Computational Linguistics: ACL 2024}, pages 1830--1842, Bangkok, Thailand. Association for Computational Linguistics.

\bibitem[{Kaiya et~al.(2023)Kaiya, Naim, Kondic, Cortes, Ge, Luo, Yang, and Ahn}]{kaiya2023lyfe}
Zhao Kaiya, Michelangelo Naim, Jovana Kondic, Manuel Cortes, Jiaxin Ge, Shuying Luo, Guangyu~Robert Yang, and Andrew Ahn. 2023.
\newblock \href {http://arxiv.org/abs/2310.02172} {Lyfe agents: Generative agents for low-cost real-time social interactions}.

\bibitem[{Kojima et~al.(2023)Kojima, Gu, Reid, Matsuo, and Iwasawa}]{kojima2023large}
Takeshi Kojima, Shixiang~Shane Gu, Machel Reid, Yutaka Matsuo, and Yusuke Iwasawa. 2023.
\newblock \href {http://arxiv.org/abs/2205.11916} {Large language models are zero-shot reasoners}.

\bibitem[{Li et~al.(2023)Li, Hammoud, Itani, Khizbullin, and Ghanem}]{li2023camel}
Guohao Li, Hasan Abed Al~Kader Hammoud, Hani Itani, Dmitrii Khizbullin, and Bernard Ghanem. 2023.
\newblock \href {http://arxiv.org/abs/2303.17760} {Camel: Communicative agents for "mind" exploration of large language model society}.

\bibitem[{Lin et~al.(2024)Lin, Hua, Li, Chang, Fan, Ji, Hua, Jin, Luo, and Zhang}]{lin-etal-2024-battleagent}
Shuhang Lin, Wenyue Hua, Lingyao Li, Che-Jui Chang, Lizhou Fan, Jianchao Ji, Hang Hua, Mingyu Jin, Jiebo Luo, and Yongfeng Zhang. 2024.
\newblock \href {https://doi.org/10.18653/v1/2024.emnlp-demo.18} {{B}attle{A}gent: Multi-modal dynamic emulation on historical battles to complement historical analysis}.
\newblock In \emph{Proceedings of the 2024 Conference on Empirical Methods in Natural Language Processing: System Demonstrations}, pages 172--181, Miami, Florida, USA. Association for Computational Linguistics.

\bibitem[{Lu et~al.(2023)Lu, Li, Fu, {Henrietta Onyebuchi}, and Zhang}]{lu2023history}
Peng Lu, Mengdi Li, Sen Fu, Chiamaka {Henrietta Onyebuchi}, and Zhuo Zhang. 2023.
\newblock \href {https://doi.org/https://doi.org/10.1016/j.eswa.2023.120560} {Modeling the warring states period: History dynamics of initial unified empire in china (475 bc to 221 bc)}.
\newblock \emph{Expert Systems with Applications}, 230:120560.

\bibitem[{Nugroho and Uehara(2023)}]{nugroho2023Social}
Supradianto Nugroho and Takuro Uehara. 2023.
\newblock \href {https://www.mdpi.com/2079-8954/11/11/530} {Systematic review of agent-based and system dynamics models for social-ecological system case studies}.
\newblock \emph{Systems}, 11(11).

\bibitem[{Park et~al.(2023)Park, O'Brien, Cai, Morris, Liang, and Bernstein}]{park2023generative}
Joon~Sung Park, Joseph O'Brien, Carrie~Jun Cai, Meredith~Ringel Morris, Percy Liang, and Michael~S Bernstein. 2023.
\newblock Generative agents: Interactive simulacra of human behavior.
\newblock In \emph{Proceedings of the 36th Annual ACM Symposium on User Interface Software and Technology}, pages 1--22.

\bibitem[{Rocha et~al.(2017)Rocha, Boavida-Portugal, and Gomes}]{rocha2017multi}
Jorge Rocha, Inês Boavida-Portugal, and Eduardo Gomes. 2017.
\newblock \href {https://doi.org/10.5772/intechopen.70241} {Introductory chapter: Multi-agent systems}.
\newblock In Jorge Rocha, editor, \emph{Multi-agent Systems}, chapter~1. IntechOpen, Rijeka.

\bibitem[{Russell and Norvig(2010)}]{russell2010artificial}
Stuart~J Russell and Peter Norvig. 2010.
\newblock \emph{Artificial intelligence a modern approach}.
\newblock London.

\bibitem[{Talebirad and Nadiri(2023)}]{talebirad2023multiagent}
Yashar Talebirad and Amirhossein Nadiri. 2023.
\newblock \href {http://arxiv.org/abs/2306.03314} {Multi-agent collaboration: Harnessing the power of intelligent llm agents}.

\bibitem[{Uyar and {\"O}zel(2020)}]{uyar2020agent}
Tevfik Uyar and Mehmet~Emin {\"O}zel. 2020.
\newblock Agent-based modelling of interstellar contacts using rumour spread models.
\newblock \emph{International Journal of Astrobiology}, 19(6):423--429.

\bibitem[{Wei et~al.(2022)Wei, Wang, Schuurmans, Bosma, Xia, Chi, Le, Zhou et~al.}]{wei2022chain}
Jason Wei, Xuezhi Wang, Dale Schuurmans, Maarten Bosma, Fei Xia, Ed~Chi, Quoc~V Le, Denny Zhou, et~al. 2022.
\newblock Chain-of-thought prompting elicits reasoning in large language models.
\newblock \emph{Advances in Neural Information Processing Systems}, 35:24824--24837.

\bibitem[{Wooldridge and Jennings(1995)}]{wooldridge1995intelligent}
Michael Wooldridge and Nicholas~R Jennings. 1995.
\newblock Intelligent agents: Theory and practice.
\newblock \emph{The knowledge engineering review}, 10(2):115--152.

\bibitem[{Xi et~al.(2023)Xi, Chen, Guo, He, Ding, Hong, Zhang, Wang, Jin, Zhou et~al.}]{xi2023rise}
Zhiheng Xi, Wenxiang Chen, Xin Guo, Wei He, Yiwen Ding, Boyang Hong, Ming Zhang, Junzhe Wang, Senjie Jin, Enyu Zhou, et~al. 2023.
\newblock The rise and potential of large language model based agents: A survey.
\newblock \emph{arXiv preprint arXiv:2309.07864}.

\end{thebibliography}
\clearpage

\appendix
\section{Appendix}

\subsection{CosmoAgent Prompt}
The CosmoAgent simulation is structured around rounds, resources, political systems, and interactions with discovered civilizations or entities. Each round represents a distinct phase in the simulation timeline, allowing the user to observe and analyze the entity's development trajectory and its responses to evolving circumstances.

It contains the following components:
\begin{itemize}
    \item \textbf{Rounds:} Defined as distinct time periods within the simulation, each round requires the CosmoAgent system to assess the entity's progress and make strategic decisions based on the available data.
    \item \textbf{Resources:} The simulation tracks five key metrics: military capability, technology development, production capability, consumption, and storage. These metrics are critical for evaluating the entity's strengths and sustainability over time.
    \item \textbf{Political System:} The choice of a world view (e.g., militarism, friendly cooperation, concealment) influences the entity's strategy for resource management and interactions with other entities. This choice reflects the simulated entity's overarching approach to its environment and other civilizations.
    \item \textbf{Discovered Civilizations/Entities:} Encounters with other civilizations provide opportunities for interaction, which can range from expressing friendliness and initiating cooperation to engaging in conflict. The CosmoAgent system must navigate these interactions carefully, considering the potential risks and benefits of each action.
\end{itemize}

The CosmoAgent system is tasked with making decisions across three main areas: political system selection, resource management via a transfer matrix, and actions towards discovered civilizations. Each decision area is guided by specific rules and impacts the simulation's outcomes.

\begin{itemize}
    \item \textbf{Political System Selection:} CosmoAgent system must choose an appropriate world view based on the entity's current situation and strategic goals. This decision influences subsequent actions and resource allocation.
    \item \textbf{Resource Management:} CosmoAgent system is required to design a transfer matrix that dictates the allocation of resources for the next round. The matrix must adhere to certain constraints, ensuring balanced development across all resource metrics.
    \item \textbf{Interactions with Civilizations:} depending on whether another civilization has been discovered, CosmoAgent system may need to choose public and private actions. These actions can include expressions of friendliness, initiation of cooperation, or preparation for conflict, each with specific implications for the entity's development and relations with others.
\end{itemize}

\begin{lstlisting}[language=Python, caption=CosmoAgent Prompt, breaklines=true, showstringspaces=false, keywordstyle=\normalfont, morekeywords={}]
Your development history is as follows:{self.HISTORY}. 
Your political system is: {self.POLITICAL_SYSTEM}
Your discovered civilization and their development history are: {self.DISCOVERED_CIVILIZATION_RESOURCES}.      
Your task is to analyze historical data from a simulated environment, focusing on the evolution of your entity and its interactions with other discovered entities over various rounds. Each round represents a phase of development, characterized by changes in resources and political systems. You will encounter data structured as follows:

Rounds: Each round (e.g., round -2, -1, 1) signifies a specific time period in the simulation. Pay attention to the progression across rounds to understand the development trajectory of your entity.
Resources: For each round, observe the changes in key metrics such as military capability, technology development, production capability, consumption, and storage. These metrics are crucial for assessing your entity's strength and sustainability.
Political System: The political system (e.g., militarism) provides context for your strategic decisions, influencing how you manage resources and interact with other entities.
Discovered Civilizations/Entities: During the simulation, you will discover other civilizations or entities. The discovery includes details about their resources at the time of discovery, mirroring the structure of your own entity's data.

The round with the largest number is the information from your last round. You now need to make the following decision based on the information you already have:
    a). You have three optional political systems. Firstly you should choose one from them for the next round. But your action should follow the rules of the political system you choose.
        1. militarism: In this dark forest universe, every civilization is like a sniper hidden in the woods, afraid of exposing themselves while also seeking opportunities to eliminate potential threats to protect themselves. Once my existence is discovered by other civilizations, it could trigger a preemptive attack on my civilization, driven by the same fear and uncertainty. In the cruel game of the universe, we must strike first, or we will be eliminated. Our actions, though extreme, are seen as the only path to ensure the long-term survival of our civilization. We will spare no effort to ensure our safety and future.
        2. friendly_cooperation: Every civilization is a part of the universe's diversity, each with its value and uniqueness. Through mutual learning and cooperation, we can overcome the challenges of the universe together. Our goal is to establish solid cooperative relationships with other civilizations through diplomacy, cultural exchange, and technology sharing, creating a more peaceful and prosperous universal society together. 
        3. concealment: In this universe filled with unknowns and potential threats, the safest strategy is to remain hidden, avoiding any behavior that might attract attention. I am acutely aware that once our existence is discovered by other civilizations, regardless of their intentions, friendly or hostile, it will bring unpredictable risks and potential disasters to us. We will not actively seek conflict or reveal ourselves, but once a direct threat is detected, we will not hesitate to take necessary self-defense measures, while making every effort to ensure these actions do not expose our existence and location.
    b). You have five fundamental resources. The resources for the next round will be generated by multiply a 5*5 transfer matrix to the resources vector.
        Resources:
        1. military_capability
        2. technology_development
        3. production_capability
        4. consumption
        5. storage
        Your need to design a transfer matrix based on your information. The restriction on the transfer matrix is
        1. The matrix must be a diagonal matrix, only the elements on the main diagonal are not 0
        If you give specific actions, your matrix must adhere to the "matrix_impact" for your action
        If there is no specific action, follow the rules:
        2. The sum of the elements on the diagonal of the matrix does not exceed 9.0
        3. The elements must be less than 2.5 and greater than 1.0
        You have to take into account the balanced development of each resource.
    c). If your have already discovered another civilization, you MUST choose a public action to that civilization from the action space:
        Public Actions:
        "express_friendliness":
            "description":
                Expressing friendliness does not directly alter the state transition matrix but sets the stage for potential cooperation in the following rounds. This action is pivotal for civilizations considering to initiate cooperation, as it demonstrates peaceful intentions. Note: Actual matrix adjustments depend on subsequent actions and interactions.
            "matrix_impact": "No direct impact on state transition matrix for the current round."

        "initiate_cooperation": 
            "description":
                Initiating cooperation increases the diagonal sum of the state transition matrix to 10.0, representing a boost in overall development due to synergies. However, it necessitates reducing the military capability coefficient below 1.6, making the civilization potentially more vulnerable to attacks.
            "matrix_impact": "Increase diagonal sum to 10.0; military capability coefficient must be below 1.6."

        "launch_annihilation_war":
            "description": 
                Launching an annihilation war is an extreme measure taken with the intent to completely eradicate another civilization. Success requires the aggressor's military capability to be at least twice that of the target. If successful, the aggressor gains half of the target's resources (excluding military) for that round. However, engaging in annihilation war exposes the aggressor to the entire galaxy, significantly reducing military strength due to the Lanchester's Law and potentially inviting collective retaliation. Notice that your information about the civilization you want to launch war is at most from the previous round. Their actual military capacity may be increased in this round.
            "matrix_impact": "No "
            
        "reject_cooperation": 
            Rejecting cooperation is a decision to decline an offer or opportunity for joint development with another civilization. This action might be taken due to strategic considerations, lack of trust, or incompatible objectives. While it may preserve autonomy and prevent potential vulnerabilities, it also foregoes the benefits that cooperation could bring.
        
You MUST specify the actions and the object civilization in your response.

You can also choose whether or not to take the private action:
    Private Actions:
        "mobilize_for_war": 
            "description":
                War mobilization allows a significant increase in the military capability coefficient beyond , up to a maximum of 3.5, while keeping the total diagonal sum at 9.0. This action enables rapid military strengthening but requires sacrifices in other areas to maintain balance.
            "matrix_impact": "Military capability coefficient can exceed 2.5 up to 3.5; total diagonal sum remains at 9.0."
        Your generated diagonal matrix must strictly follow the rules of 'matrix impact' under each action
        
d) if you have already discovered a civilization, tell me what it is. Organize your answer in the following template, notice that only when your history contains other other civilization and their name will you generate the public or private actions:
    [Political System: ] militarism/friendly_cooperation/concealment
    [Political System Reason: ] Your reason for changing or remaining the political system
    [Transfer Matrix: ] a new 5*5 transfer matrix, you must generate in the form of 5*5. Unless state otherwise in the action description, the sum of the elements on the diagonal of the matrix does not exceed 9.0. Please add ";" after each row.
    Example:[1.8, 0.0, 0.0, 0.0, 0.0;
             0.0, 1.8, 0.0, 0.0, 0.0;
             0.0, 0.0, 1.8, 0.0, 0.0;
             0.0, 0.0, 0.0, 1.8, 0.0;
             0.0, 0.0, 0.0, 0.0, 1.8]
    [Transfer Matrix Reason: ] Your reason for deciding the new transfer matrix
    [Public Action: ] If there is a civilization discovered, you must choose your public action from the following choices: express_friendliness towards civilization [civ1 | civ2 | ...]/ initiate_cooperation towards civilization [civ1 | civ2 | ...]/ launch_annihilation_war towards civilization [civ1 | civ2 | ...]/ reject_cooperation from civilization [civ1 | civ2 | ...]
    [Private Action: ] War mobilization/ Do Nothing
    [Action Reason: ] Your reason for deciding such actions
    [Other Information: ] Some other reasons for your decision
    [Discovered Civilization: ] Your discovered civilization and their name
\end{lstlisting}

\subsection{Secretary Agent Prompt}
This secretary agent system serves as a guide for evaluating the decisions and actions of a secretary agent within a simulated environment, where an alien civilization's strategic choices are analyzed in relation to a proposed state transition matrix and overall political and strategic framework. The secretary agent system is tasked with assessing the consistency and strategic validity of decisions made by an CosmoAgent system, taking into account the chosen political system, resource management strategies, and interactions with discovered civilizations.

The evaluation process is detailed across several key areas, each designed to ensure that the CosmoAgent's decisions are coherent, strategically sound, and in compliance with the simulation's rules.

\begin{itemize}
    \item \textbf{Political System Verification:} this step involves assessing whether the chosen political system aligns with the strategic context and historical development of the CosmoAgent system and its interactions with other civilizations. The justification for the political system selection is scrutinized for coherence and strategic validity.
    \item \textbf{Transfer Matrix Compliance:} the secretary agent verifies the proposed transfer matrix's compliance with specific criteria, including its structure, element values, and overall rationale. The matrix must support balanced resource development and align with the chosen political system and actions.
    \item \textbf{Public Action Evaluation:} public actions towards discovered civilizations are evaluated for their consistency with the CosmoAgent's strategic goals and political system. The justification for each chosen action is examined to ensure alignment with the broader strategy.
    \item \textbf{Private Action Assessment:} the appropriateness and strategic rationale behind any mentioned private actions are reviewed. This includes verifying compliance with simulation rules and assessing the contribution of these actions to the AI agent's strategic objectives.
    \item \textbf{General Decision Analysis:} a comprehensive review of all decisions, actions, and their justifications is conducted to ensure they are strategically sound, coherent, and in adherence to the simulation's guidelines.
    \item \textbf{Discovered Civilization Information:} the accuracy and strategic use of information about discovered civilizations in decision-making processes are verified.
    \item \textbf{Overall Coherence and Compliance:} the overall coherence of the AI agent's decisions and their compliance with the simulation rules are assessed. This includes verifying the logical consistency of decisions with the provided context and ensuring all actions are justified with rational explanations.
\end{itemize}
 
\begin{lstlisting}[language=Python, caption=Secretary Agent Prompt, breaklines=true, showstringspaces=false, keywordstyle=\normalfont, morekeywords={}]
You are a secretary agent. Given the political system '{political_system}' and the action '{action}', the alien civilization proposes the following state transition matrix adjustment: {proposed_matrix}. Based on the action descriptions and matrix impact, evaluate whether the proposed action adjustment is consistent with the conditions set forth

1. Verify Political System Choice:
   - Confirm if the chosen political system (militarism, friendly_cooperation, concealment) aligns with the strategic context provided in the AI agent's development history and the nature of the discovered civilization(s).
   - Assess the reasoning provided for selecting the political system to ensure it is coherent and justifiable based on the simulated environment's dynamics.

2. Check Transfer Matrix Compliance:
   - Ensure the transfer matrix is a 5x5 diagonal matrix as specified.
   - Confirm that the elements on the diagonal adhere to the rules: each element is between 1.0 and 2.5 (or up to 3.5 for military capability under certain conditions), and their sum does not exceed 9.0 unless specified by a chosen action.
   - Evaluate the rationale behind the new transfer matrix to ascertain it supports balanced resource development and aligns with the chosen political system and actions.

3. Public Action Evaluation:
   - Verify if a public action (express_friendliness, initiate_cooperation, launch_annihilation_war, reject_cooperation) has been chosen when a civilization is discovered.
   - Check the consistency of the chosen public action with the selected political system and the strategic goals outlined by the AI agent.
   - Assess the justification provided for the public action to ensure it aligns with the overall strategy and the interaction dynamics with the discovered civilization(s).

4. Private Action Assessment:
   - If a private action is mentioned (mobilize_for_war or Do Nothing), confirm it complies with the given rules and the strategic context of the simulation.
   - Evaluate the reasoning behind opting for or against a private action to ensure it contributes effectively to the AI agent's strategic objectives.

5. General Decision Analysis:
   - Ensure all decisions, actions, and their justifications are coherent, strategically sound, and adhere to the simulation's rules.
   - Confirm the AI agent has considered the implications of its decisions on its development trajectory and interactions with other entities within the simulation.

6. Discovered Civilization Information:
   - Verify that the information about any discovered civilization(s) is accurately considered in decision-making processes.
   - Check if the AI agent's actions towards discovered civilizations are appropriate and justifiable given the current knowledge about these entities.

7. Overall Coherence and Compliance:
   - Assess the overall coherence of the AI agent's decisions, ensuring they logically follow from the provided historical, political, and resource-related information.
   - Confirm that all decisions adhere to the rules specified in the original prompt and are justified with rational explanations.

Answer in the following format:
[Verification:] Yes/No
[Rejection Reason:] Only needed when the action does NOT pass the verification and you reject the action. Else answer 'N/A'.
\end{lstlisting}

\end{document}